\documentclass[10pt,conference,letterpaper]{IEEEtran}
\IEEEoverridecommandlockouts

\usepackage[utf8]{inputenc}
\usepackage{cite}
\usepackage{amsmath}
\usepackage{amssymb}
\usepackage{graphicx}
\usepackage{textcomp}
\usepackage{xcolor}
\usepackage{float}
\usepackage{algorithm}
\usepackage{algpseudocode}
\usepackage{booktabs}

\def\BibTeX{{\rm B\kern-.05em{\sc i\kern-.025em b}\kern-.08em
    T\kern-.1667em\lower.7ex\hbox{E}\kern-.125emX}}

\title{Unified Deep Learning Platform for Dust and Fault Diagnosis in Solar Panels Using Thermal and Visual Imaging}

\author{
    \IEEEauthorblockN{Abishek Karthik\IEEEauthorrefmark{1}, Sreya Mynampati\IEEEauthorrefmark{1}, Pandiyaraju V\IEEEauthorrefmark{1}}
    \IEEEauthorblockA{\IEEEauthorrefmark{1}Department of Computer Science and Engineering,\\
    School of Computer Science and Engineering,\\
    Vellore Institute of Technology, Chennai, India}
}

\begin{document}

\maketitle

\begin{abstract}
Solar energy is one of the most abundant and tapped sources of renewable energies with enormous future potential. Solar panel output can vary widely with factors like intensity, temperature, dirt, debris and so on affecting it. We have implemented a model on detecting dust and fault on solar panels. These two applications are centralized as a single-platform and can be utilized for routine-maintenance and any other checks. These are checked against various parameters such as power output, sinusoidal wave (I-V component of solar cell), voltage across each solar cell and others. Firstly, we filter and preprocess the obtained images using gamma removal and Gaussian filtering methods alongside some predefined processes like normalization. The first application is to detect whether a solar cell is dusty or not based on various pre-determined metrics like shadowing, leaf, droppings, air pollution and from other human activities to extent of fine-granular solar modules. The other one is detecting faults and other such occurrences on solar panels like faults, cracks, cell malfunction using thermal imaging application. This centralized platform can be vital since solar panels have different efficiency across different geography (air and heat affect) and can also be utilized for small-scale house requirements to large-scale solar farm sustentation effectively. It incorporates CNN, ResNet models that with self-attention mechanisms-KerNet model which are used for classification and results in a fine-tuned system that detects dust or any fault occurring. Thus, this multi-application model proves to be efficient and optimized in detecting dust and faults on solar panels. We use different metrics like Precision, accuracy, F1 score and a lot more to track the performance of this novel classification system. We have performed various comparisons and findings that demonstrates that our model has better efficiency and accuracy results overall than existing models.

\end{abstract}

\section{Introduction}

Electricity, an indispensible fuel for economic growth and daily needs; is used across every aspects of life and can be obtained from various sources. The traditional sources are generally non-renewable energy sources like coal and petroleum which is not only limited but also harms the environment with excessive tapping and usage.

Today, the sense of accountability to introduce renewable energy sources could not be more imperative. While the dreadful problem of climate change, and the urgency to lessen the emission of greenhouse gases, is close, going to renewable energy resources is imperative. The main traditional fossil fuel sources being backbone of global energy production, besides primary contributing into the level of carbon emissions they also degrade environment and air worsening the pollution level. Renewable energy such as solar, wind, hydroelectric and geothermal are sources that have low impact and do not leave any pollution. Without limiting on climatic challenges, renewable energy sources are the ones that accelerate people's energy self-sufficiency, reduce contingencies, as well as create job opportunities and promote investment in green technologies. Transitioning to renewable source of energy is a lot more than just ecological responsibility. Digging deep, it gives us the opportunity to create a more protected, intrinsic and enduring future for the generations to come.

With vast abundance, scalability and versatility-solar energy is the best green energy alternative. Additionally, advancements in technology have made solar panels more efficient and affordable, driving widespread adoption and reducing reliance on fossil fuels. With minimal emissions and clean environmental impact, it promises a better and safe environment and growth for our future generations.

The efficient operation of solar panels plays a crucial role in harnessing Solar energy and ensuring Sustainable power generation. However, various factors such as faults and dust accumulation can significantly degrade the performances of solar panels. Therefore, the accurate and finely faults and dust on solar panels is essential for optimal functioning and maintenance.

We have come up with a novel approach for fault detection and dust detection on solar panels using thermal image dataset and DL techniques. Thermal images provide valuable information about the temperature distribution on the panel surface, which can be indicative of faults or dust accumulation. DL algorithm, specifically Residual Networks (ResNet) have shown remarkable performance in image analysis tasks.

\begin{figure}[htbp]
    \centering
    \includegraphics[width=0.8\linewidth]{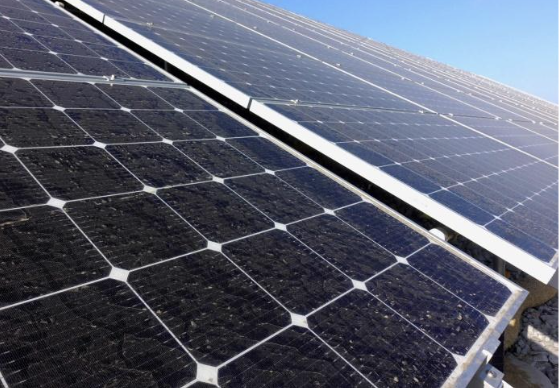} 
    \caption{Solar panel to generate renewable energy commercially}
    \label{fig:solar_commercial}
\end{figure}

\begin{figure}[htbp]
    \centering
    \includegraphics[width=0.8\linewidth]{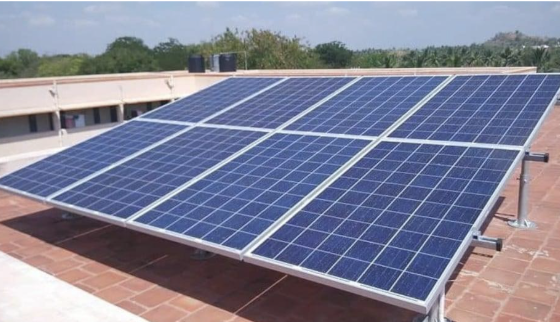} 
    \caption{A solar panel installed for domestic household use}
    \label{fig:solar_domestic}
\end{figure}

\begin{figure}[htbp]
    \centering
    \includegraphics[width=0.8\linewidth]{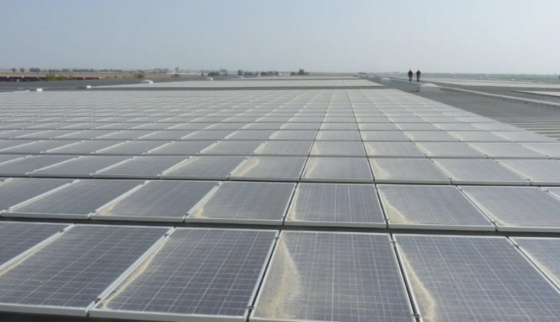} 
    \caption{Dirty solar panel pictorial representation}
    \label{fig:dirty_panel}

\end{figure}\begin{figure}[htbp]
    \centering
    \includegraphics[width=0.8\linewidth]{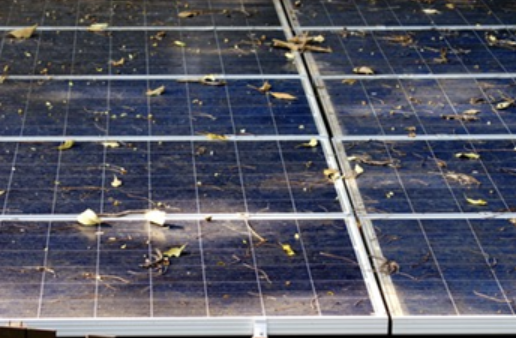} 
    \caption{A solar panel covered with leaf remains (dirty solar panel)}
    \label{fig:leaf_panel}
\end{figure}

\section{Literature Review}

In 2020, Danial Saqiab et al.~\cite{ref1} had implemented a proposal that draws a stark contrast between theoretical efficiency (80\%) and practical efficiency output (15-17\%). They list out various reasons on why a solar panel or cell module produce such different outputs than calculated like geography, heat, dust and various other factors that impact the cell output. The paper focuses on irradiance being input and has threshold limit voltage set while sensor is used to predict voltage in the panel. When voltage drops below threshold it can be used to trigger the system to clean the panel and maintain it as such it functions with high performance. Thus, a voltage limit trigger unit is set that can be used to result for maintenance and cleaning based on the limit set as per required.

In 2022, Saif Hassan Onim and et al.~\cite{ref2} have collected datasets throughout Bangladesh and have trained existing models such as AlexNet, CNN, ResNet. They have proposed a SolNet model that performs better than all previous mentioned models (98.2\%). SolNet has provided better results with less parameters which reduces computational complexity and time to train; so we can say it performs better since it provides accuracy with lesser complexity and time-taken (the model is not real-time and has some offset difference). However, model over-fit is a common barrier to producing robust models and testing the models with different variations of data is the way to detect it. The authors have used K-fold variations and early stopping in the proposed SolNet to overcome this barrier. The SolNet architecture of CNN model has been developed in such a way that it can provide better efficiency with accuracy under better performance and time that comparing with other SOTA models.

In 2019, Ali Unluturk and et al.~\cite{ref3} have proposed a paper that tells us about the different environmental conditions have a substantial impact on solar photovoltaic (PV) power plants' energy yield. The temperature of the PV modules, dust, shade, and solar irradiation are a few of them that stand out. In order to examine how the dust shading factor affects energy efficiency, power outputs for three different densities of dust collection on the module surface are evaluated in this study using artificial light in a laboratory setting. Based on the Gray Level Co-occurrence Matrix, new features are derived from the PV module images captured by a camera for varying degrees of dust build-up. The PV module performance is then obtained by Artificial Neural Networks that that classifies obtained data to get impact dust level and provides with accurate result.

In 2023, El karch hajar and et al.~\cite{ref4} have proposed YOLO-a CNN based model that performs better than R-CNN model in their work. The authors have come up with various parameters that can be used to evaluate and detect soiling and has addressed with solution for effective evaluation of soil detection and also has automatic methods for soil and dust detection. They have used a PV system including a standalone PV module of four panels of Poly-crystalline silicon and automatic technologies for the model with effective evaluation strategies. They have also verified performance against artificial light to check against sunlight and dust affect to the PV module in its operating since different regions across globe have different factors affecting them. The paper has shown an overall accuracy of 80\% output in detecting dust and sand over panels from different dataset images and classifications. It has implemented and processed using YOLOv5 framework for improvising outcome and computations.

In 2020, Jianyu Wang and et al.~\cite{ref5} have proposed a paper that vividly describes what is soiling and subsequent degradation of performance of PV modules. The paper has focused on sinusoidal echo network (SENs) over traditional Infrared thermography of shaded PV cells for finding shaded cell performance and its degradation. They have utilized YOLO v3, Resnet, Mask R-CNN models for their project. The proposed abnormal shading detect system is divided into three steps. The first two steps use YOLOv3, and the object recognition part uses Mask R-CNN. First use labelling to label the field images manually. In order to improve the robustness of the model, only train the yolov3 model with the labelled PV panel images at this stage, then extract and process the photovoltaic panel images collected in the field, at last sets the portions of the detected panels as RoI (region of interest) (using FCN model) and proceed to the next step (YOLOv3) to locate the abnormal shading. According to the testing findings, our system has a classification accuracy rate of more than 94\%.

In 2017, Dirk.C.Jordan and et al.~\cite{ref6} have come forward with a system to find PV cell deterioration using clear-sky irradiance and year-on-year rate estimates instead of traditional on-site sensors. They have derived estimates that robustly calculates deterioration despite soiling, sensor drift and other factors. The cleansing is done after data is filtered and cleansed using clear-sky filter (csi); degradation analysis done. The evaluation metrics held by them are like data shift and patterns of dataset, variable of calibration state of irradiation sensor, soiling and cloudy climate and its changes and finally other nonlinearities that can impact irradiance and shading resulting in affect of PV functioning. Their application includes PV fleet roll-ups and validating NREL PV systems using above methods.

In 2020, Mingda Yang and et al.~\cite{ref7} introduced a paperwork that investigates into the effectiveness of image-based techniques for quantifying soiling on solar panels, with a focus on understanding how various imaging conditions impact the accuracy of results. The study systematically evaluates factors such as lighting conditions, camera settings, and image processing algorithms to determine their influence on soiling quantification accuracy. The method's viability is evaluated by creating a prototype image processing system and lab setting. More precisely, images of a surrogate surface with controlled dust loadings are used to compute the black/white ratio. The final findings display that the possibility of using aerial photos for PV panel soiling quantification in the future, as dust loading may be measured under a variety of suitable imaging situations.

In 2023, Po Ching Hwang and et al.~\cite{ref8} have come up with a model that tells us the significance of of the shadow that soiling casts, its effects that may eventually lead to a decrease in the production of electricity. The classification models are used to determine whether or not the panels require cleaning by receiving the RGB images. Numerous image processing algorithms, including edge detection, histogram analysis, filtering, and color masks, are also used in these techniques. An IV-curve of soiling impact and voltage is also drawn. The spread of soiling is assessed by combining a statistical technique with an AI system. Ultimately, the decision system receives both the soiling rate and soiling distribution to decide when and if the PV modules should be cleaned. The use deep learning and statistics to determine whether or not the soiling distribution is uniform in order to address the nonuniform soling problem.

In 2023, Amir Sohail and et al.~\cite{ref9} the authors speak about how PV cells have become a mainstream source of energy harvest and how it has helped large industries and economies to scale. The suggested work takes two key techniques to solve these approaches' drawbacks. Initially, a powerful deep-learning technique is suggested to identify the many kinds of PV cell cracks, including deep and microcracks. The orientation of the crack in a microcrack determines its classification, which is important. The power analysis is then carried out in accordance with the degree of cracking. They have incorporated models using four ML models namely-U-net, LinkNet, FPN, and attention U-net are trained, assessed, and compared for the purpose of crack identification. Several metrics are used to assess the models' efficiency, such as intersection over union (IoU) and F1-Score. Through the use of neural networks, the study offers a robust and automated solution for identifying and mitigating faults in photovoltaic cells, thereby increasing the reliability and efficiency of solar power generation.

In 2016, Hassan Qasem and et al.~\cite{ref10} the author tells us about the barriers and various other spheres like irradiation which affects significantly the functioning of PV modules. They use high-definition cameras on drones called aerial robots to collect images and check them. The work advances techniques for monitoring and controlling the performance of photovoltaic systems by demonstrating the effectiveness of the suggested approach in precisely detecting dust levels on solar panels through validation through experimentation and comparative analysis.

In 2022 Krishna Sandeep Ayyagiri and et al.~\cite{ref11} have proposed a paper speaks about how PV technology and solar panels can be conjoined and used among other sources of renewable energy generating plants to obtain green energy. It describes how the emissions from these plants can affect or pollute the solar panels and how the dust accumulated over time can be degrading the optimal performance of these panels. The proposed method takes into account PV array images captured by a high-resolution camera. They have used CNN classification methods for feature extraction and training and have made use of Long short term memory (LSTM) models for learning sequential meteorological data to enable joint classification and detection tasks to reduce computational complexities. The accuracy for CNN is 51\% and LSTM has 84\% while combined CNN-LSTM has accuracy as high as 94.09\% proving to be robust and effective.

In 2023, Ghada Shaban Eldeghady and et al.~\cite{ref12} have come forward with a model about significant impact caused by various PV system failures that can arise and directly impair system performance. One of the many contributions of this work is the analysis of PV performance in different failure scenarios utilizing fault detection features including maximum power, voltage at maximum power point, open circuit voltage, and short circuit current. The diagnostic technique's execution time and speed are decreased when this method is combined with other works. The back propagation neural network and PSO technique performance of the proposed algorithm is evaluated in terms of PV system problem diagnosis. This method combines the local search power of the back propagation neural network with the global search power of the PSO algorithm. The BPNN-PSO techniques improve the convergence of the diagnostic procedure and increase the forecast accuracy of fault diagnostics for solar systems. The primary addition is the amalgamation of a deep learning neural network approach and a heuristic optimization strategy, which can offer an enhanced learning process for accurately predicting problems.

In 2022 Rawad Al-Mashhadani and et al.~\cite{ref13} proposes an important step to improve the sustainability and efficiency of solar energy systems is to explore the potential of deep learning in detecting solar PV panel failures using infrared solar panel photographs. Twelve classes (cell, cell mulch, crack, diode, diode mulch, hotspot, hotspot mulch, no anomaly, offline module, shading, pollution, vegetation) were represented in the 20,000 images of the dataset modules used in this study. The Efficient0 model exemplar was utilized in the process. This paper examines the most significant studies that have recently used deep learning to analyze solar energy failures. They have emphasized the importance of both deep and hybrid learning models and listed out its advantages and differences of each separately for the people to understand comprehend well.

In 2023 Ragul.S and et al.~\cite{ref14} this model explores the integration of cloud computing and machine learning techniques for the detection of electrical faults in photovoltaic (PV) systems. By mixing the scalability and accessibility of cloud resources with predictive power of ML models, it tries to enhance the efficiency in finding PV panel faults. The necessary data for creating the algorithm is obtained by simulating the photovoltaic system in the MATLAB/Simulink environment under different operating conditions. The proposed model is validated through experiments, achieving 100\% training accuracy and 97.399\% testing accuracy using a randomly divided dataset. The study illustrates the potential of this integrated approach to enhance the performance and reliability of PV systems by enabling proactive maintenance and fault mitigation techniques through thorough analysis and experimentation.

In 2023, Fuhao Sun and et al.~\cite{ref15} have introduced us to a model which explores and identifies all kinds of existing machine learning work to identify solar cell dust detection and reviews them in-depth. The article then discusses contemporary training strategies, such as the use of machine learning algorithms to train our system and reduce laboriousness and improve accuracy of the dust detection process while enhancing robustness and efficiency. It highlights about the present problems in developing and maintaining solar panels (like high dust in gulf regions) to directions of future trends. Concluding, this paper provides a helpful overview of current research on dust detection techniques for solar panels and points to intriguing future developments in this field.

In 2023 Kumwar Baba Ali and et al.~\cite{ref16} proposed a work that explores the different ways of cleaning solar photovoltaic cells to bring about development in green energy sphere globally. It explores the spheres of smart and innovative technology like 5G and beyond to help aid in detecting dust and cleaning them periodically. The proposed system provides a thorough analysis of residue sorting strategies for PV systems that leverage the combined power of ML and the Internet of Things. The authors have assessed the effectiveness and productivity of IoT and ML-based ways for reducing residue arrangement on solar-powered chargers by examining previous research and tactics.

In 2022, Siyuan Fan and et al.~\cite{ref17} have implemented a work to detect dust accumulation on photovoltaic cells using a DRNN (deep residual neural network) to find dust concentration. Then, to classify the dust accumulation, an image preprocessing method is designed that uses transformation, clustering, nonlinear interpolation, equivalent segmentation. The \(R^2\) and mean absolute error of the DRNN are 78.7\% and 3.67, respectively. Also, the authors have designed three methods to identify uneven dust accumulation on cells. Thus, the above approach can help PV systems operate and maintain themselves intelligently.

In 2022 Benjamin Oluwamuyiwa Olorunfemi and et al.~\cite{ref18} the authors look into using smart systems to monitor and wipe the dirt off solar panels in order to improve performance. The review analyzes several approaches and technology used for dirt detection and cleansing in solar power systems by synthesizing previous research. By lessening the effects of dirt accumulation, the authors examine the efficacy and efficiency of smart systems, which include automation, data analytics, and sensors, in enhancing solar panel performance. The report attempts to maximize energy output and sustainability by offering observations on the present state of research and proposing future possibilities for the development of intelligent solutions for solar panel cleaning and monitoring through this hierarchial and monitored review.

In 2022, S.Prabhakaran and et al.~\cite{ref19} the authors have tried to optimize the existing PV cell cleaning methods and introducing Real-Time Multi Variant Deep learning Model (RMVDM) to enhance performance in the article. The approach detects and localizes flaws by considering and identifying several factors such as dust, micro-cracks, cracks, and spotlights. Region-Based Histogram Approximation (RHA) algorithm has been incorporated for image processing module. The preprocessed images are applied with Gray Scale Quantization Algorithm (GSQA) to extract the features. With above mentioned MVDM model, extracted features are trained using a number of layers, each belonging to different classes of neurons. Each class neuron has been designed to measure Defect Class Support (DCS). The technique locates the flaw in the image and determines its class by using the output layer's return of several DCS values. Additionally, the fault is localized using the HigherOrder Texture Localization (HOTL) technique using the method. The proposed technique and deep learning model produces optimized results with around 97\% in fault detection and localization with better accuracy and shorter time period.

In 2020, Ritu Maity and et al.~\cite{ref20} in this review, the authors describe a unique approach that uses convolutional neural networks (CNNs) to forecast power loss in order to identify dust on solar panels. To precisely detect and measure the amount of dust present, the method uses CNNs to analyze RGB dust photos taken from solar panels. The scientists want to offer a trustworthy way to evaluate the effect of dust deposition on solar panel performance by establishing a correlation between predicted power loss and dust levels. By enabling proactive maintenance and cleaning tactics based on real-time dust detection and power loss prediction, the chapter advances ways for maximizing solar energy production through innovation.

In 2020 Ayush Bishta and et al.~\cite{ref21} proposed a paper examining the methods used today in solar photovoltaic (PV) panel cleansing systems reveals a wide range of strategies used by the sector. In order to increase energy output and minimize performance deterioration, this survey emphasizes the significance of effective maintenance techniques. Furthermore, research into how machine learning (ML) technology could be incorporated into these systems in the future shows promise for improved automation, predictive maintenance, and optimization. Proactive cleaning scheduling, problem detection, and performance optimization may be achieved to a large extent by using machine learning (ML) methods, such as neural networks and predictive analytics, to evaluate sensor data, weather predictions, and historical performance. To fully reap the benefits of machine learning in solar photovoltaics, however, issues including data integrity, complexity of model, and set up costs need to be properly addressed.

In 2020, Joshua Arockia Dhanraj and et al.~\cite{ref22} have made a study that evaluates in-depth a number of techniques used to locate defects in solar panels, from conventional methods to cutting-edge innovations like artificial intelligence and machine learning. Through a comprehensive analysis of extant literature and practical data, the writer provides valuable perspectives on the efficiency, dependability, and constraints of various defect detection methodologies. The study also covers the significance of prompt defect detection for maintaining peak performance, optimizing energy output, and extending solar panel life. This study adds to the current discussion on defect detection in solar energy systems by means of a thorough examination and analysis, offering practitioners, researchers, and stakeholders in the renewable energy industry useful direction.

In 2018, A.Syafiq and et al.~\cite{ref23} have come up with a paper that provides a thorough summary of the most recent advancements in solar photovoltaic (PV) panel self-cleaning technology. This study explores several strategies and tactics that are meant to improve solar panel performance and efficiency by using self-cleaning mechanisms. Through their analysis of current developments in the fields of materials research, surface engineering, and automation technologies, the writers offer insightful perspectives on the possible advantages and difficulties of integrating self-cleaning systems into solar energy generation. In addition, the study addresses how these developments may affect environmental sustainability, cost-effectiveness, and scalability, which will aid in the continuous endeavor to maximize solar PV installation upkeep and operation.

\subsection{Research Gap}

Fault discovery of solar panels in solar granges is a critical aspect of icing the trustability and performance of these systems. Blights in solar panels can lead to a number of problems that affect the overall effectiveness and cost-effectiveness of solar granges. These issues include reduced performance, increased conservation costs, safety pitfalls and reduced panel life. One of the most significant consequences of imperfect solar panels is reduced affair. When a solar panel malfunctions, it produces lower electricity than a healthy panel. This power reduction can accumulate throughout the solar ranch, eventually reducing overall energy product. Due to the significant investment needed for solar granges, any reduction in performance has direct profitable consequences, which is why early discovery and form of panel faults is essential.

In addition to reduced performance, defective solar panels can also lead to increased conservation costs. Regular conservation is necessary to insure optimal operation of solar granges. Still, imperfect panels frequently bear more frequent and technical attention, adding overall conservation costs. These increased costs burden the solar ranch driver and can impact the profitability of the renewable energy system. Safety pitfalls are another problem associated with defective solar panels. Similar panels can emit heat or sparks, posing a implicit threat to workers who are responsible for servicing and maintaining the solar ranch. These safety hazards can lead to accidents and injuries, further emphasizing the significance of prompt fault discovery and remediation to alleviate these pitfalls. In addition, the reduced lifetime of imperfect solar panels adds to the overall operating costs of a solar ranch. A failed panel may not last as long as a healthy one, taking it to be replaced sooner than anticipated. This early relief demand increases the cost associated with purchasing and installing new panels, further affecting the profitable viability of a solar ranch.

Addressing these issues is essential not only to maximize energy performance, but also to minimize the operating and conservation costs associated with solar power installations. The primary challenge is to develop an effective and dependable system for detecting and diagnosing solar panel failures and for monitoring and controlling dust and dirt accumulation. Current error discovery styles frequently calculate on regular homemade checks or introductory detector grounded approaches. These traditional styles may not give timely and accurate information about the status of individual panels or the entire system. Homemade checks are labor ferocious and may miss crimes that aren't incontinently visible. Introductory detector-grounded approaches may be limited in their capability to identify nuanced issues. In addition, the accumulation of dust on solar panels can lead to significant energy losses. Dust and debris on the face of solar panels act as a hedge, reducing the quantum of sun that can be absorbed and converted into electricity. Current cleaning ways, frequently homemade or semi-automatic, are generally hamstrung and precious. They can also introduce the threat of damaging the panel during cleaning, further complicating the operation and conservation of the solar ranch. Thus, there's an critical need for more effective and accurate fault discovery and drawing styles for solar granges. The perpetration of advanced technologies similar as machine literacy, IoT and drones can significantly ameliorate disfigurement discovery and the cleaning process. Machine literacy algorithms can dissect data from colourful detectors to identify faults and prognosticate conservation requirements. IoT bias can give real-time monitoring and reporting, enabling visionary conservation. Drones equipped with cameras and detectors can check the panels and clean them effectively while reducing security pitfalls.

It is very tedious and difficult to obtain various images for dataset and training since there are no proper sources and networks for obtaining solar panel images. Moreover, there is no proper standard of classification of dataset images like soiling, shadowing, fault, crevices etc based on the defect and this makes it irrelevant to work on without classifying on own.

The cost for implementing and integrating this system on real-time with proper standards and maintenance would be too high and hard to operate. Since, there are a lot of factors that affect or degrade performance it is difficult to find and modify for any defects detected.

The geography and environment of these PV panels also widely varies and can differently impact its utilization performance. This means that countries like in middle east with abundant sunlight are prone to more dust by nature and have more defects and soiling etc that ultimately impacts optimization and this varies globally by geography.

In conclusion, early discovery and form of solar panel faults are consummate to optimizing solar ranch performance, safety and frugality. Advanced technologies and innovative approaches are necessary in working the problems associated with fault discovery and cleaning, icing that solar energy remains a sustainable and effective source of renewable energy.

\subsection{Contributions of Work}

The perfect functioning and operating of solar panels is an integral part of the success of solar energy systems. However, malfunctions and dust accumulation on solar panels can significantly hinder their performance. This chapter addresses the critical need for accurate and timely detection of such problems and proposes a solution that combines thermal imaging and deep learning techniques for defect and dust detection.

Solar panel failures include various types, including electrical, mechanical, and manufacturing defects. These failures can lead to reduced energy performance, increased maintenance costs and even complete system failure. On the other hand, the accumulation of dust gradually covers the surfaces of the panels, reducing their ability to effectively absorb solar radiation and endangering the energy conversion.

To address these issues, this study utilizes the potential of thermal imaging, which provides insight into the temperature distribution on panel surfaces. Deep learning, specifically Resnet, is used to analyze thermal images and detect faults and dust patterns. This new approach is in line with a broader effort in solar energy technology to improve monitoring and maintenance capabilities.

We would be making an intelligent fault detection system that would help us to detect solar cell malfunctioning, faults, crevices and defects on the PV panel using real-time sensors.

Investigate the accumulation of dust and other particles on the surfaces of solar panels by examining state-of-the-art detecting systems like optical and electrostatic sensors.

Make a robust diagnostic system that would recommend real-time suggestions based on severity of faults and actions that would impact minimal downtime and optimal performance.

Make a better inference from understanding the relation between the amount of dust accumulation to the extent of power loss and then use this correlation to maintain optimal performance.

Combining the fault and dust detection techniques under a centralized platform so as to be user-friendly platform for all to test and maintain their panels individually against standard operational measures.

Perform comprehensive field testing and experiments in actual solar panel installations to verify the designed defect detection and dust monitoring system's accuracy, dependability, and efficacy.

Now, perform analysis based on our findings and conclusions to make energy savings model, cost-to-benefit chart and other recommendations like environmental benefits for our centralized fault/dust detection model.

Contribute to the technological development of renewable energy: You may encourage information sharing and expand the framework for sustainable energy solutions by publishing research findings, techniques, and technological developments in prestigious publications and conferences.

\section{Overall Proposed System Architecture}

Maintaining the efficiency and durability of photovoltaic systems depends on dust detection and solar panel failure. Deep learning and thermal imaging are two tools used in problem solving that help detect and categorize different types of issues. Temperature anomalies that signify failure can be found using thermal imaging, which records temperature changes brought on by flaws or dust buildup. Deep learning methods like convolutional neural networks (CNN) and residual networks (RESnet) are trained to discern between problems and events by processing these images. Solar panel operators can strengthen the stability of the solar system, solve faults, improve performance, and lower maintenance costs by using this inspection procedure. Since thermal imaging can detect light problems that are invisible to the human eye, it offers a non-invasive and efficient method of keeping an eye on the condition of solar panels. Deep learning models can be adjusted to fit various fault and dust model types after being trained on diverse datasets. The benefits of this strategy include enhanced capacity and emergency maintenance, both of which are critical for sizable solar arrays. Operators can avoid power outages and price reductions by including this technology into the regular maintenance of solar panels. This allows for the prompt detection and correction of issues. Ultimately, solar photovoltaic systems become more dependable and efficient when thermal imaging and deep learning are combined, increasing their effectiveness and affordability.

The foundation of residual learning is the hypothesis that neural networks ought to pick up residual functions, or the differences between the input and the intended output. Training gets harder and harder in classic deep networks as the number of layers rises because of issues like vanishing gradients. ResNet is innovative because it makes use of residual blocks, which enable the construction of extraordinarily deep networks without compromising performance. Skip connections, often referred to as shortcut connections, are among these leftover blocks. They circumvent one or more levels, enabling the network to ``skip'' specific layers and go straight to the activations of the preceding layer. Skipped connections make it easier for gradients to flow during training, solve the issue of disappearing gradients, and make deep network optimization easier. ResNet architectures are hence scalable to hundreds or even thousands of layers without sacrificing their excellent accuracy during training and testing.

In the realm of deep learning, ResNet has established itself as a mainstay, making substantial contributions to the fields of object identification, computer vision, picture classification, and more. His architectural breakthrough not only made it easier to train deeper networks, but it also established the benchmark for convolutional neural network design in the contemporary era. ResNet's impact is still noticeable in a number of situations where deep learning is essential.

We have also used gamma removal and Gaussian filtering mechanisms among other similar preprocessing functions like resizing, normalization and much more. Now, to improve accuracy and efficiency of our model, we have incorporated the use of self-attention mechanisms to these existing models. We have used kernet architecture into ResNet to bring about better and accurate results. This proposed system brings about a centralized platform for both dust and fault detection among solar panels and is at it robust and efficiently.

\begin{figure}[htbp]
    \centering
    \includegraphics[width=0.8\linewidth]{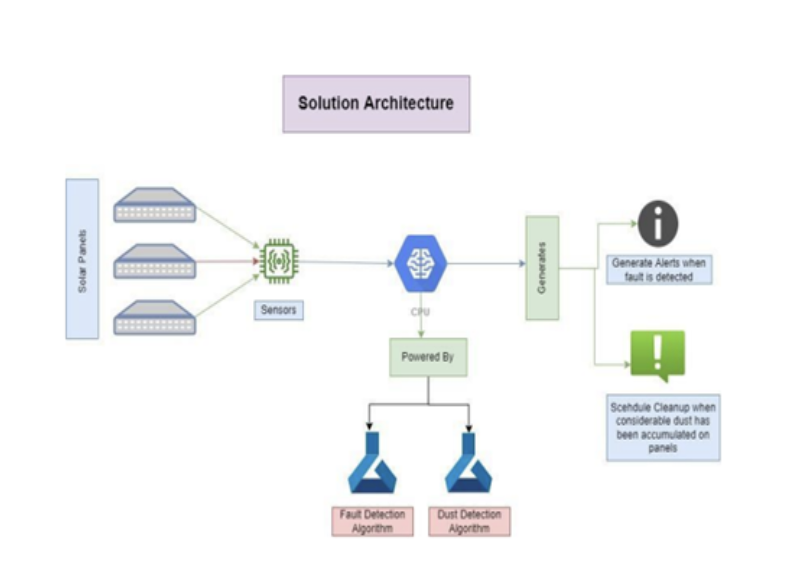} 
    \caption{Solution architecture pictorial representation}
    \label{fig:solution_arch}
\end{figure}

\begin{figure}[htbp]
    \centering
    \includegraphics[width=0.8\linewidth]{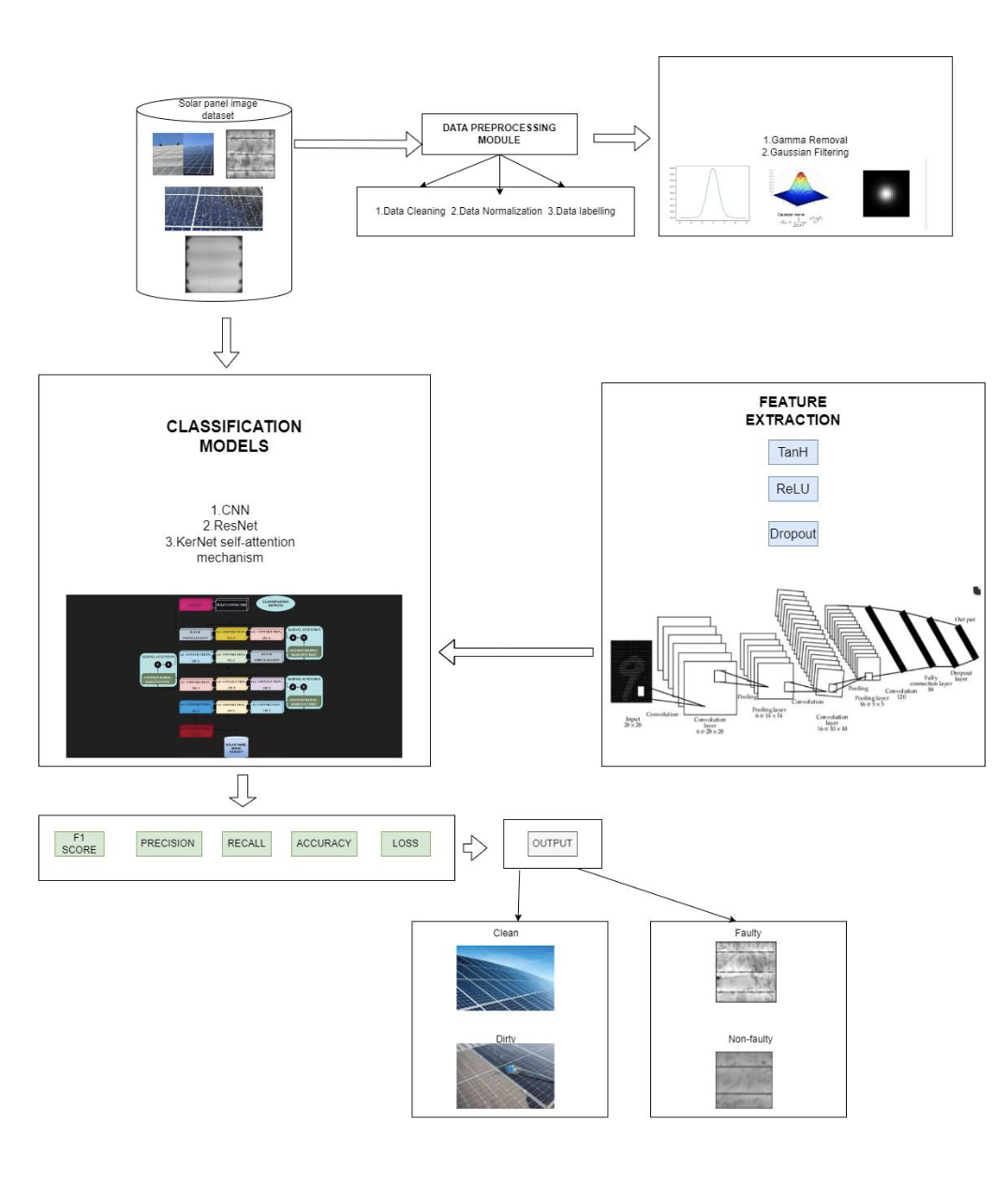} 
    \caption{Architectural Diagram of proposed system}
    \label{fig:arch_diagram}
\end{figure}

\subsection{Dataset Exploration}

Both our applications have two dataset classifications each such as dirty/clean and faulty/non faulty dataset for various solar panel images. The total application consists of 2654 images of different solar panels and is used to test and train our model to improvise and increase efficiency and robustness.

The first application of dust detection on solar panels uses dirty/clean datasets to classify solar panel images respectively and produces result classifying it under either of these. The next application uses faulty/non-faulty dataset, visualized to classify the solar panel image as faulty or well-functioning based on different parameters which we will see later.

\subsubsection{Dust Detection}

Dataset consists of dusty/dirty and clean solar panel images classified into separate folders for training and testing purposes. The images are of varying sizes and conditions captured under different lighting and environmental scenarios.

\subsubsection{Fault Detection}

Input Dataset comprises thermal images of solar panels with various fault types and non-faulty panels. Sample images are preprocessed and labeled as faulty or non-faulty based on manual inspection and standard criteria.

\begin{figure}[htbp]
    \centering
    \includegraphics[width=0.8\linewidth]{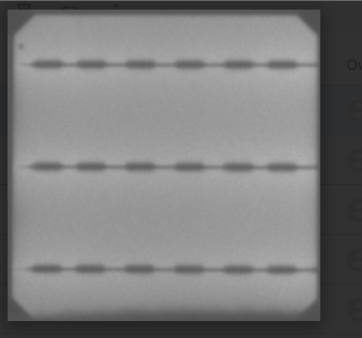}
    \caption{Sample image from dataset}
    \label{fig:sample_dataset}
\end{figure}

\subsection{Preprocessing}

While doing the data acquisition we will be working in two categories; our first category will be to use the thermal camera images done in the frame of asset maintenance and the second one will be done to find out the faults in the solar panels. The convolution and the act of gamma adjustment was performed during the preprocess procedure which was done in order to have a precise dataset.

Use pre-processing methods under transformation functions and gamma removal, gaussian filtering as well to classify data from both obtained datasets to obtain proper results.

We deploy two types of algorithms to further increase robustness and efficiency of preprocessing module namely gamma removal and Gaussian filtering.

\begin{figure}[htbp]
    \centering
    \includegraphics[width=0.8\linewidth]{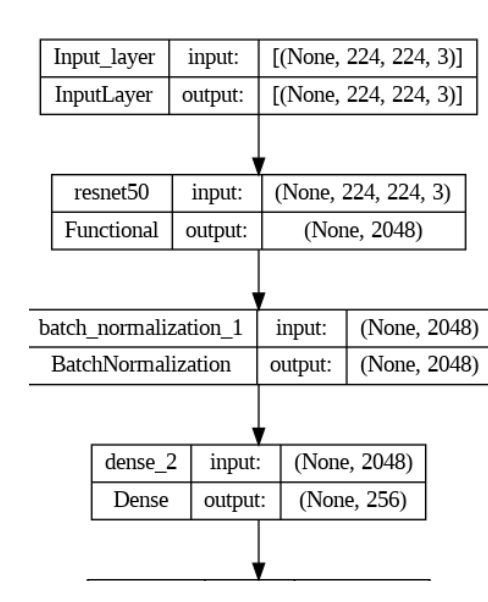}
    \caption{Preprocessing workflow diagram}
    \label{fig:preprocess_workflow}
\end{figure}

\begin{figure}[htbp]
    \centering
    \includegraphics[width=0.8\linewidth]{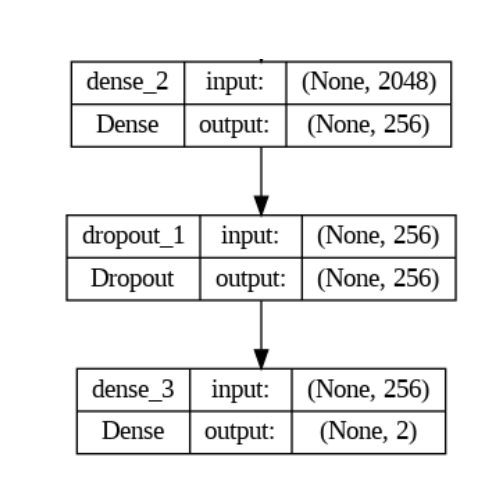}
    \caption{In-built functionalities like ReLU and Dropout used in ResNet}
    \label{fig:relu_dropout}
\end{figure}

\begin{figure}[htbp]
    \centering
    \includegraphics[width=0.8\linewidth]{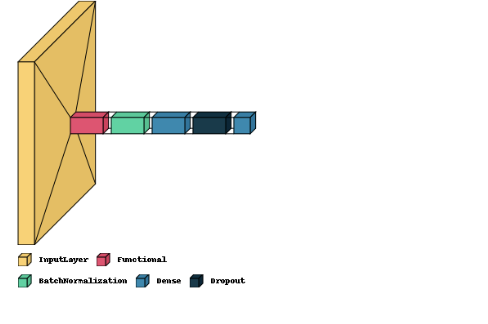}
    \caption{Gaussian filtering process illustration}
    \label{fig:gaussian_filter}
\end{figure}

\subsubsection{Gamma Removal}

Gamma removal, often referred to as gamma correction, is a non-linear operation used to adjust the brightness and contrast of an image. The operation involves applying an inverse gamma function to the pixel values of an image. The gamma function is defined as:

\[
G(x) = x^{\frac{1}{\gamma}}
\]

\subsubsection{Gaussian Filtering}

\textbf{ResNet Block with Gaussian Filtering}

Let \(x\) be the input to the ResNet block, \(F(x)\) be the output, and \(H(x)\) be the function that the residual block approximates.

The output of the ResNet block with Gaussian filtering is given by:

\[
F(x) = H(x \ast G) + x
\]

Where \(\ast\) represents the convolution operation, and \(G\) is the Gaussian kernel.

By substituting the expressions for convolution and the Gaussian kernel, we get:

\[
F(x) = \sum_{m}^{} \sum_{n}^{} x(m,n) \cdot G(i - m, j - n, \sigma) + x
\]

Where \((i,j)\) represents the spatial position of the output.

This expression represents the output of the ResNet block with Gaussian filtering, where the input \(x\) is convolved with the Gaussian kernel \(G\) to produce the filtered result, which is then added back to the input.

\textbf{Algorithm 1: Solar Panel Image Preprocessing}

\begin{algorithmic}
\State \textbf{input:} raw solar panel images
\State \textbf{output:} preprocessed solar panel images
\Function{preprocessImages}{$img_{solarPanel}$}
    \State $\delta \leftarrow$ gamma\_factor
    \For{each pixel $P$ in $img_{solarPanel}$}
        \State $P \leftarrow P^{\delta}$
    \EndFor
    \State $k \leftarrow$ kernel size
    \State $g(i,j) \leftarrow$ create gaussian kernel
    \State Normalize the gaussian kernel:
    \[
    g_{norm}(i,j) \leftarrow \frac{g(i,j)}{\sum_{i=-k}^{k} \sum_{j=-k}^{k} g(i,j)}
    \]
    \State $filtered\_image \leftarrow$ convolve$(img_{solarPanel}, g_{norm})$
    \State where:
    \[
    \text{convolve}(x,y) \leftarrow \sum_{i=-k}^{k} \sum_{j=-k}^{k} img_{solarPanel}(x-i, y-j) \cdot g_{norm}(i,j)
    \]
    \State $img_{preprocessed} \leftarrow filtered\_image$
    \State \textbf{return} $img_{preprocessed}$
\EndFunction
\end{algorithmic}

\begin{figure}[htbp]
    \centering
    \includegraphics[width=0.8\linewidth]{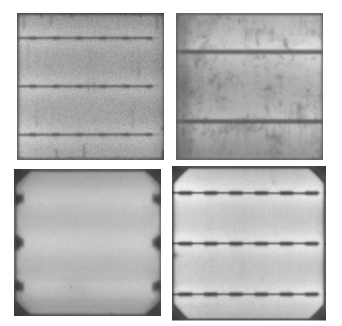}
    \caption{Output of fault detection module after preprocessing.}
    \label{fig:preprocess_output}
\end{figure}

\subsection{Classification}

The data would be processed, solutions would be determined for a number of faults and chances would be greater for the machine to break down less with help of choosing algorithms such as Real-Time environmental networking (R-Net) that includes a self-attention mechanism to fully train data and receive various ratings on which to base the correct results for the purpose.

The model, which will be fed by both the test group as well as the data and train sets specified for each particular outcome (to detect the magnitude of the abnormality and do fault detection or no) will generate a final decision. An accurate structure is to incorporate the same configuration for both values, which will achieve the required result.

Use ResNet for improvising model and other parameters by benefitting from use of residual blocks and much more layers. Added self-attention mechanism like kernet to it to improvise overall model.

A mathematical expression for a ResNet block incorporating convolutional layers followed by max pooling or average pooling, let's start with the basic ResNet block structure:

1. \textbf{Input:} The \(x\) is the function's input block.
2. \textbf{Output:} \(F(x)\) constituting output of the block.
3. \textbf{Residual Mapping:} Residual block\((x)\) is the function that \(H(x)\) tries to fit.

The output of the ResNet block is given by:

\[
F(x) = H(x) + x
\]

Now, let's consider integrating different types of layers into the ResNet block:

\textbf{A ResNet Block with a Convolutional Layer.}

We will denote Conv\((x)\) as what comes out of a convolutional layer when the input is \(x\). So, the residual mapping \(H(x)\) with a convolutional layer can be represented as:

\[
H(x) = \text{Conv}(x)
\]

And thus, the output of the ResNet block with a convolutional layer is:

\[
F(x) = \text{Conv}(x) + x
\]

\textbf{ResBlock with Max Pooling Layer} is also another method we have implemented to solve this issue. Here, we let MaxPool\((x)\) be the output of the pooling layer, which is called the max pooling layer, by the matrix \(x\). So, the residual mapping \(H(x)\) with a max pooling layer can be represented as:

\[
H(x) = \text{MaxPool}(\text{Conv}(x))
\]

And thus, the output of the ResNet block with a max pooling layer is:

\[
F(x) = \text{MaxPool}(\text{Conv}(x)) + x
\]

\textbf{Blocks with ResNet has the Average Pooling Layer.}

Let's assume AvgPool\((x)\) is the output of the convolutional layer using average pooling algorithm for the input \(x\). So, the residual mapping \(H(x)\) with an average pooling layer can be represented as:

\[
H(x) = \text{AvgPool}(\text{Conv}(x))
\]

And thus, the output of the ResNet block with an average pooling layer is:

\[
F(x) = \text{AvgPool}(\text{Conv}(x)) + x
\]

For instance, factors of people with bunch of layers being inserted in ResNet block without interfering Residual learning element. The convolutional layer looks for complex features, but the pooling layers handle fewer dimensions and still get the other important patterns.

\subsubsection{KerNet Model}

By enabling each feature map to concentrate on pertinent spatial areas, kernel attention mechanism model is applied to ResNet to improve its feature extraction. This improves the network's flexibility and performance in tasks such as segmentation and image identification.

Let's denote \(x\) as the input to the ResNet block, \(F(x)\) as the output, and \(H(x)\) as the function that the residual block approximates. Additionally, let \(A(x)\) represent the attention mechanism applied to the input \(x\).

The output of the ResNet block with kernel attention can be formulated as:

\[
F(x) = H(A(x)) + x
\]

Here's a more detailed derivation:

\textbf{Attention Mechanism:} The attention mechanism \(A(x)\) assigns attention weights to different spatial locations in the input feature maps. Let's denote \(A(x)\) as the attention-weighted representation of the input \(x\).

\textbf{Residual Mapping:} The residual mapping \(H(x)\) operates on the attention-weighted input \(A(x)\) to produce the residual features to be added to the input. This mapping might involve one or more convolutional layers to capture spatial dependencies.

Putting it all together, the output of the ResNet block with kernel attention is the sum of the residual mapping applied to the attention-weighted input and the original input:

\[
F(x) = H(A(x)) + x
\]

\subsubsection{Batch Normalisation}

The activations of each layer to stabilise the training process is normalised.

The equation for Batch Normalisation:

\[
\text{output} = \text{BatchNormalization}(\text{Conv}(\text{input}))
\]

\[
Y = \frac{(X - \mu)}{\sqrt{\sigma^2 + \varepsilon}} \cdot \gamma + \beta
\]

Where:
\begin{itemize}
    \item \(X\) is the input tensor
    \item \(Y\) is the output tensor after batch normalisation
    \item \(\mu\) is the mean of the batch
    \item \(\sigma\) is the variance of the batch
    \item \(\gamma\) is a learnable scaling parameter
    \item \(\beta\) is a learnable shifting parameter
\end{itemize}

\subsubsection{Max-Pooling Layer}

Max pooling decreases spatial dimensions of the feature maps and retains important information. It reduces computational difficulty. It is represented as:

\[
\text{output} = \text{MaxPooling}(\text{input}, \text{pool\_size}, \text{strides}, \text{padding})
\]

\[
N(\text{output}) = \left\lfloor \frac{N(\text{input}) + 2p - k}{s} \right\rfloor + 1
\]

Where:
\begin{itemize}
    \item pool\_size: size of the pooling window
    \item strides: the step size
    \item padding: `valid' or `same'
\end{itemize}

\subsubsection{Convolutional Layer}

Given an input tensor \((X)\) of size \((H \times W \times C_{in})\), where \((H)\) is the height of the input, \((W)\) is the width of the input, and \((C_{in})\) is the number of input channels, and a filter \((F)\) of size \((K \times K \times C_{in})\), where \((K)\) is the height and width of the filter, the convolution operation at a specific spatial location \((h, w)\) in the output tensor can be calculated as:

\[
Y(h,w,c) = \sum_{i=0}^{K-1} \sum_{j=0}^{K-1} \sum_{d=0}^{C_{in}-1} X(h+i, w+j, d) \cdot F(i, j, d, c)
\]

Where:
\begin{itemize}
    \item \(Y(h, w, c)\) is the output tensor value at the spatial location \((h, w)\) in the output tensor for the filter number \((c)\).
    \item \(X(h + i, w + j, d)\) is the input tensor value at spatial location \((h + i, w + j)\) in the input tensor for input channel \((d)\).
    \item \(F(i, j, d, c)\) is the filter value at spatial location \((i, j)\) in the filter for input channel \((d)\) and filter number \((c)\).
\end{itemize}

\begin{figure}[htbp]
    \centering
    \includegraphics[width=0.8\linewidth]{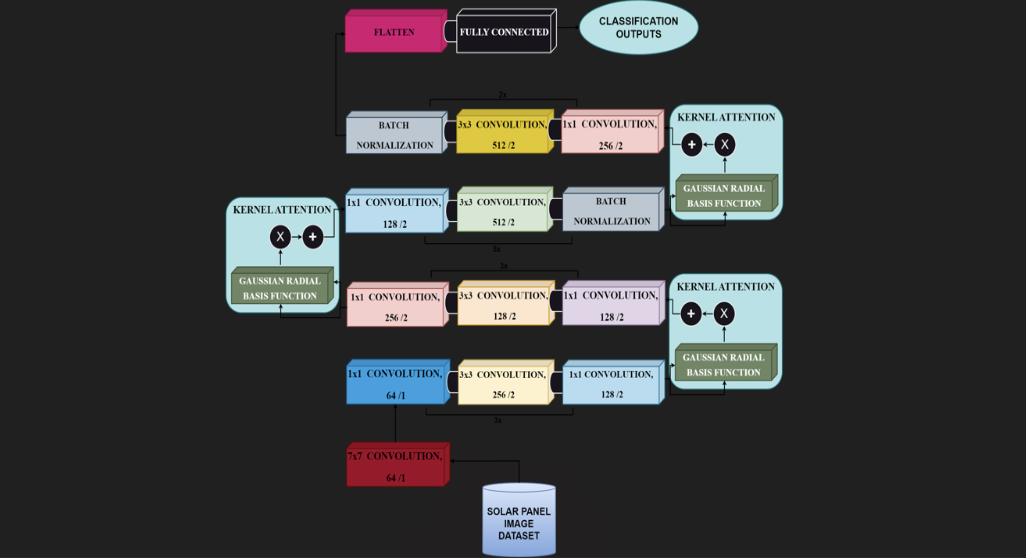}
    \caption{Layer-Architectural diagram of the model}
    \label{fig:layer_arch}
\end{figure}

\textbf{Algorithm 2: KerNet Training for Fault and Dust Detection}

\begin{algorithmic}
\State \textbf{input:} preprocessed solar panel images
\State \textbf{output:} trained KerNet classifier for fault and dust detection
\Function{train\_kerNet}{$img_{solarPanel}$}
    \State $b \leftarrow$ batch size
    \For{each epoch}
        \State $\text{batch} \leftarrow$ fetch the next batch of images
        \State $\text{probs} \leftarrow$ Compute predicted class probabilities using KerNet on batch
        \State calculate binary crossentropy loss, $\delta$ between predictions and labels:
        \[
        L(y, \hat{y}) \leftarrow -\frac{1}{N} \sum_{i=1}^{N} [y_i \cdot \log(\hat{y_i}) + (1-y_i) \cdot \log(1-\hat{y_i})]
        \]
        \State update model parameters $\theta$ using backpropagation with loss $\delta$
        \[
        \theta' \leftarrow \theta - \mu \nabla \delta
        \]
        \State where $\nabla \delta$ is the gradient of loss $\delta$ with respect to model parameters $\theta$
        \State compute accuracy:
        \[
        \text{accuracy} = \frac{TP + TN}{TP + TN + FP + FN}
        \]
        \State compute precision:
        \[
        \text{precision} = \frac{TP}{TP + FP}
        \]
        \State compute recall:
        \[
        \text{recall} = \frac{TP}{TP + FN}
        \]
        \State compute F1 Score:
        \[
        \text{F1-Score} = \frac{2 \cdot TP}{TP + TN + FP}
        \]
        \State where:
        \begin{itemize}
            \item $TP \leftarrow$ true positives
            \item $TN \leftarrow$ true negatives
            \item $FP \leftarrow$ false positives
            \item $FN \leftarrow$ false negatives
        \end{itemize}
        \If{loss increases}
            \State update learning rate to further learning
        \EndIf
        \State save model weights
    \EndFor
    \State \textbf{return} trained model
\EndFunction
\end{algorithmic}

\section{Results and Discussions}

This paper provides an intuitive approach and a centralized platform for us to detect dust and fault on photovoltaic cells. The combined platform with dual application can be easily adaptable and used for any purposes ranging from small homes to large solar farms.

The model has been trained to give very accurate and quick results for both of these applications and works robust and efficiently. This work has proposed a centralized platform which can be modified and developed further easily in future. It can be further enhanced in future by combining with innovations like satellite imagery so as to improvise it to work even better.

\subsection{Output Screenshots}

The input solar panel image data are classified as clean/dirty as output prediction. The model predicts whether a solar panel (image) is faulty or non-defective using training methods and provides output with accuracy.

\begin{figure}[htbp]
    \centering
    \includegraphics[width=0.8\linewidth]{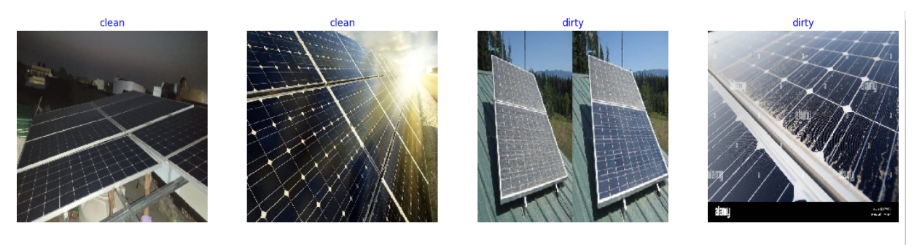} 
    \caption{Input solar panel image data are classified as clean/dirty as output prediction.}
    \label{fig:output_classification}
\end{figure}

\subsection{Performance Evaluation}

\subsubsection{Configuration and Environment Parameters}

\begin{table}[H]
\centering
\begin{tabular}{ll}
\toprule
\textbf{Parameter} & \textbf{Value} \\
\midrule
Operating System & Microsoft Windows 11 Home \\
CPU & Intel(R) Core(TM) i5-1035G1 CPU @ 1.00GHz, 1201 MHz \\
Number of Cores & 4 \\
GPU & Tesla-P100-16GB \\
CUDA & Version 11.8 \\
\bottomrule
\end{tabular}
\end{table}

\subsubsection{Parameter Specifications}

\begin{table}[H]
\centering
\begin{tabular}{ll}
\toprule
\textbf{Parameters} & \textbf{Coefficients} \\
\midrule
Total number of layers & 224 \\
Learning Rate & 0.001 \\
Epochs & 130 \\
Batch Size & 32 \\
\bottomrule
\end{tabular}
\end{table}

We have developed a deep learning system where the model classifies solar panels as either dusty (positive class) or non-dusty (negative class), the confusion matrix is a 2x2 matrix that includes:

The performance of the classification model can be evaluated using various metrics derived from the confusion matrix:

\textbf{Accuracy:} The proportion of correctly classified solar panels (both dusty and non-dusty):

\[
\text{Accuracy} = \frac{TP + TN}{FP + FN + TP + TN}
\]

\textbf{Precision:} The proportion of correctly classified dusty panels among those classified as dusty:

\[
\text{Precision} = \frac{TP}{FP + TP}
\]

\textbf{Recall (Sensitivity or True Positive Rate):} The proportion of actual dusty panels correctly classified as dusty:

\[
\text{Recall} = \frac{TP}{FN + TP}
\]

\textbf{Specificity:} The proportion of actual non-dusty panels correctly classified as non-dusty:

\[
\text{Specificity} = \frac{TN}{FP + TN}
\]

\textbf{F1-Score:} The harmonic mean of Precision and Recall, providing a balance between the two metrics:

\[
\text{F1-Score} = \frac{2 \cdot (\text{Precision} \times \text{Recall})}{\text{Precision} + \text{Recall}}
\]

\begin{figure}[htbp]
    \centering
    \includegraphics[width=0.8\linewidth]{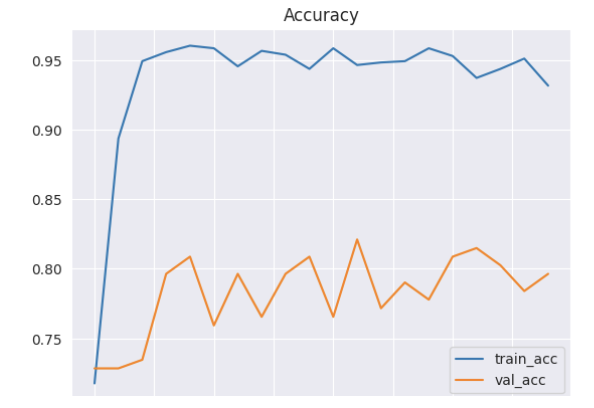}
    \caption{Training accuracy over epochs}
    \label{fig:train_accuracy}
\end{figure}

\begin{figure}[htbp]
    \centering
    \includegraphics[width=0.8\linewidth]{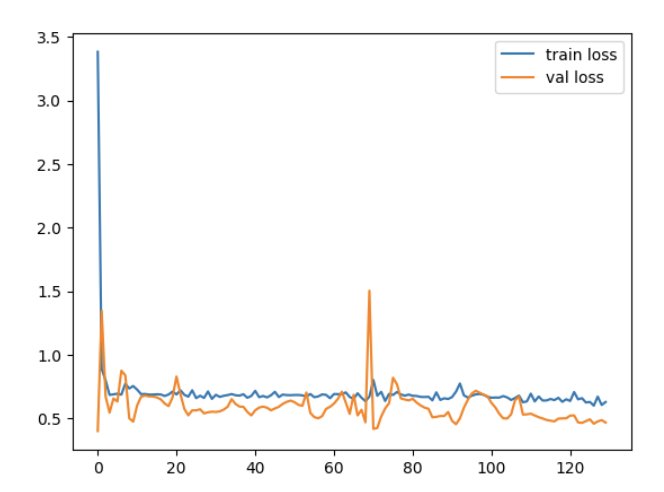}
    \caption{Training loss over epochs}
    \label{fig:train_loss}
\end{figure}

\begin{figure}[htbp]
    \centering
    \includegraphics[width=0.8\linewidth]{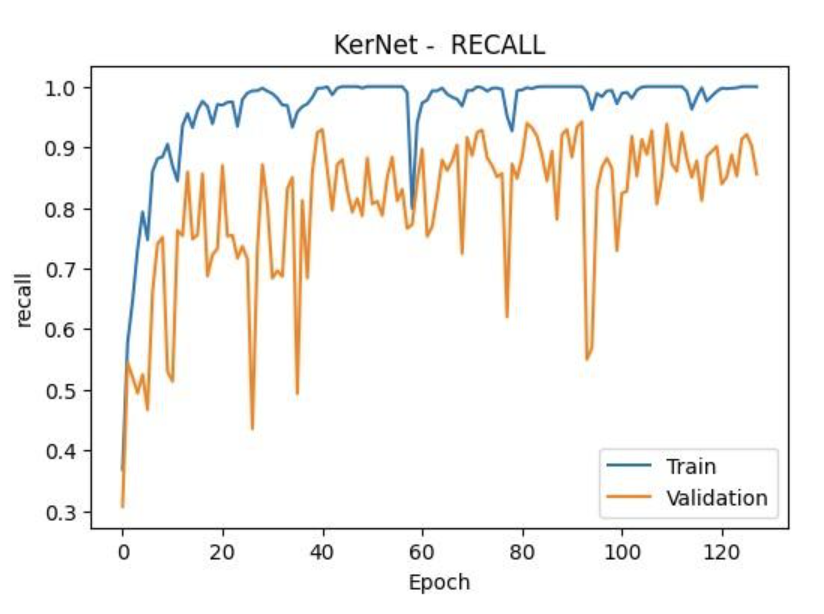}
    \caption{Validation accuracy over epochs}
    \label{fig:val_accuracy}
\end{figure}

\begin{figure}[htbp]
    \centering
    \includegraphics[width=0.8\linewidth]{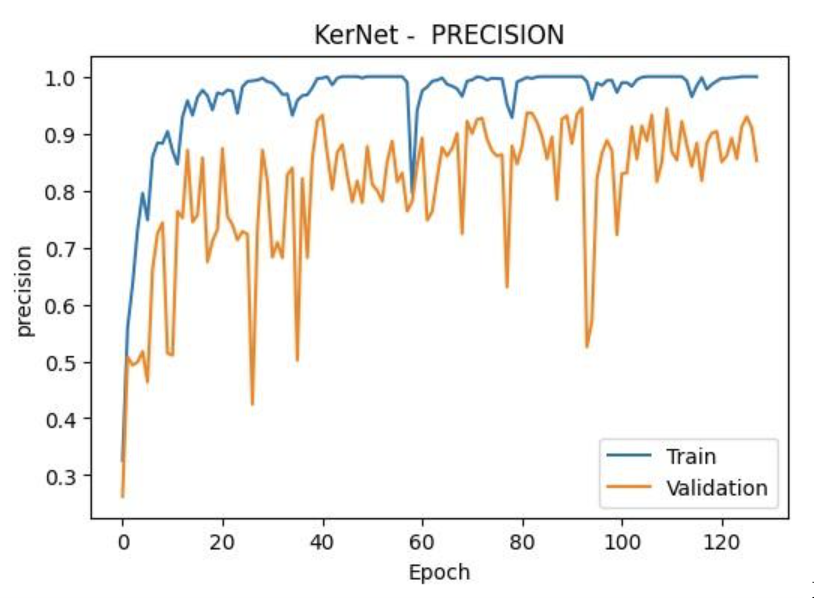}
    \caption{Validation loss over epochs}
    \label{fig:val_loss}
\end{figure}

\begin{figure}[htbp]
    \centering
    \includegraphics[width=0.8\linewidth]{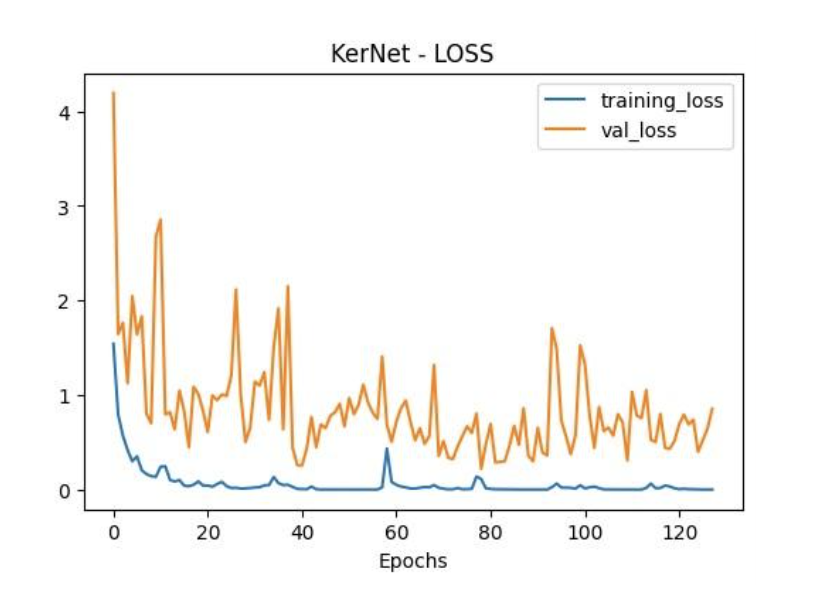}
    \caption{Precision metric over epochs}
    \label{fig:precision_metric}
\end{figure}

\begin{figure}[htbp]
    \centering
    \includegraphics[width=0.8\linewidth]{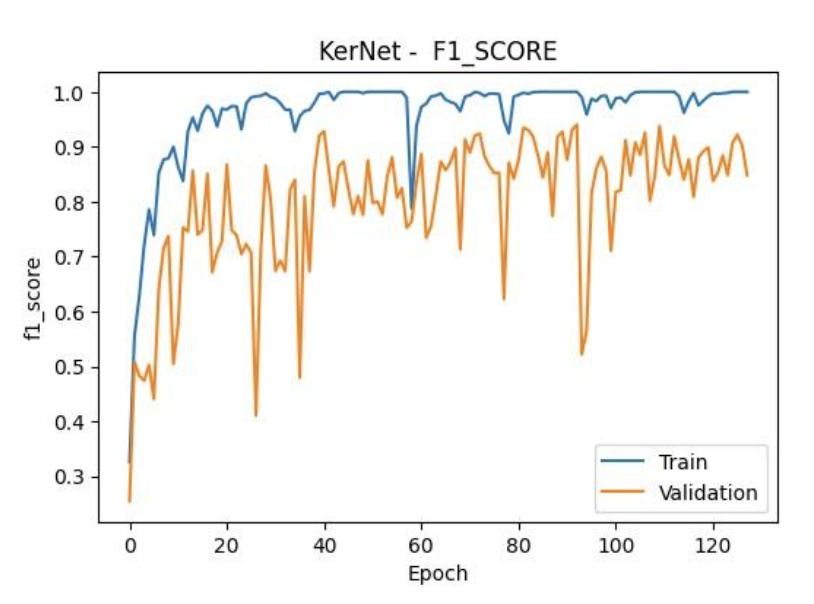}
    \caption{Recall metric over epochs}
    \label{fig:recall_metric}
\end{figure}

\begin{figure}[htbp]
    \centering
    \includegraphics[width=0.8\linewidth]{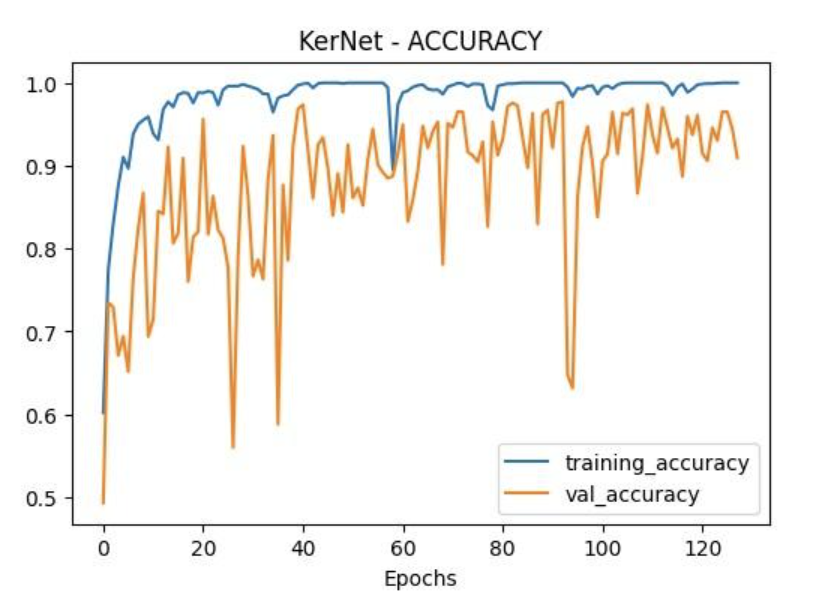}
    \caption{F1-Score metric over epochs}
    \label{fig:f1_metric}
\end{figure}

\subsection{Confusion Matrix}

True Positive (TP): The number of solar panels that are actually dusty and the model correctly classified them as dusty.

True Negative (TN): The number of solar panels that are actually non-dusty and the model correctly classified them as non-dusty.

False Positive (FP): The number of solar panels that are actually non-dusty but the model incorrectly classified them as dusty. This is also known as a Type I error.

False Negative (FN): The number of solar panels that are actually dusty but the model incorrectly classified them as non-dusty. This is also known as a Type II error.

The confusion matrix can be represented as follows:

\begin{table}[H]
\centering
\begin{tabular}{lll}
\toprule
& \textbf{Actual Dusty} & \textbf{Actual Non-Dusty} \\
\midrule
\textbf{Predicted Dusty} & TP & FP \\
\textbf{Predicted Non-Dusty} & FN & TN \\
\bottomrule
\end{tabular}
\end{table}

\begin{figure}[htbp]
    \centering
    \includegraphics[width=0.8\linewidth]{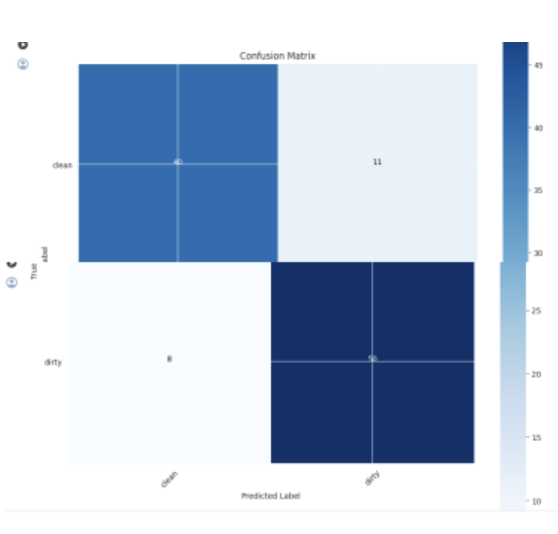}
    \caption{Confusion matrix for the proposed model}
    \label{fig:confusion_matrix}
\end{figure}

\subsection{Comparative Analysis}

\begin{table}[H]
\centering
\begin{tabular}{lllll}
\toprule
\textbf{Parameters} & \textbf{VGG16} & \textbf{AlexNet} & \textbf{SOTA} & \textbf{Proposed KerNet} \\
\midrule
F1-Score & 0.6 & 0.73 & 0.85 & 0.99 \\
Sensitivity & 0.9 & 0.86 & 0.8 & 0.99 \\
Specificity & 0.92 & 0.82 & 0.94 & 0.99 \\
Precision & 0.93 & 0.9 & 0.97 & 0.99 \\
Accuracy & 0.96 & 0.94 & 0.98 & 0.99 \\
\bottomrule
\end{tabular}
\caption{Comparative Analysis of different proposed models for solar panel dust and fault detection application}
\end{table}

\begin{figure}[htbp]
    \centering
    \includegraphics[width=0.8\linewidth]{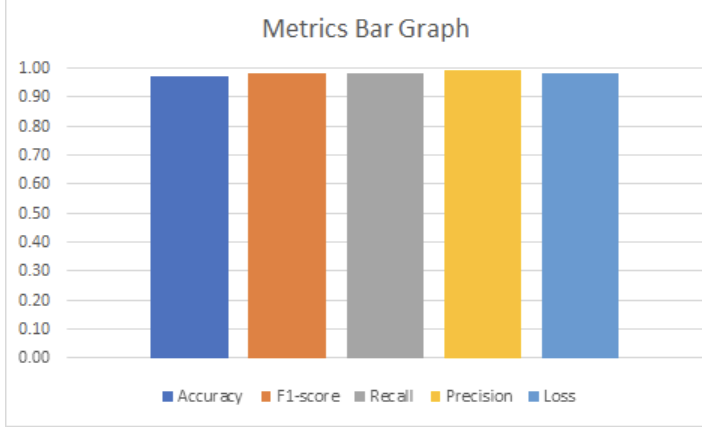}
    \caption{Comparative analysis of F1-Score across models}
    \label{fig:compare_f1}
\end{figure}

\begin{figure}[htbp]
    \centering
    \includegraphics[width=0.8\linewidth]{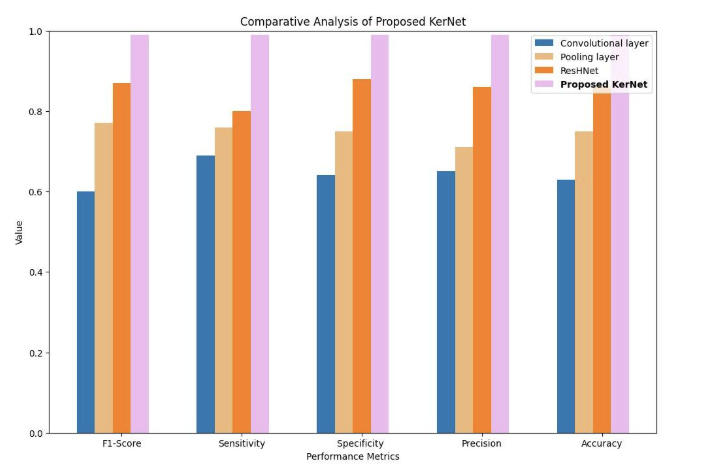}
    \caption{Comparative analysis of accuracy across models}
    \label{fig:compare_accuracy}
\end{figure}

\begin{figure}[htbp]
    \centering
    \includegraphics[width=0.8\linewidth]{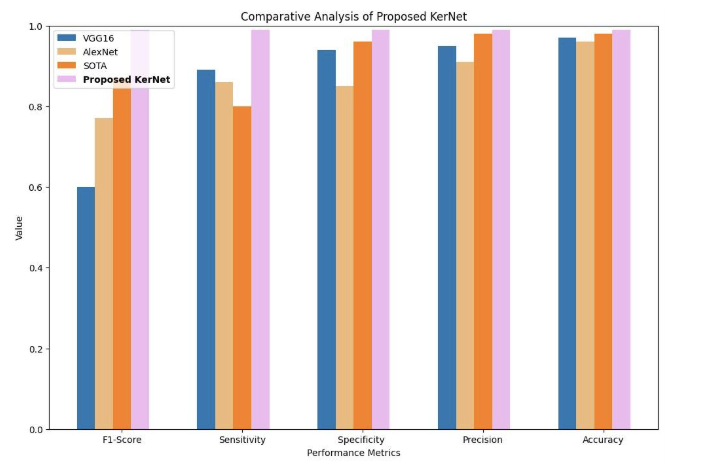}
    \caption{Comparative analysis of precision and recall across models}
    \label{fig:compare_prec_recall}
\end{figure}

\subsubsection{Comparison With Existing Work}

Solar filter models use layers to study images for dust. These layers may miss key differences in image parts. Now your model can do better. ResNet can catch details at different levels. It helps spot dust. This means the habituation mechanism is best. They're like guards in the layers. The way they see images is not fixed. They can focus on spots with signs like color or texture changes.

This way has two big pluses. One, it can boost accuracy by focusing on key spots. Also, the kernel attention works well without costing too much time. It means short training and fast off-line work.

Still, the real model might beat a random one in just some cases. But ResNet and kernel focus might top all. They could be more exact and work faster for real-time use.

However, the real model will still beat the random model only in the very specific cases when it is used in the same conditions as this random model. Nevertheless, the synergy of ResNet's deep feature learning and the focused attention based on kernels can have capability of overpassing current models, and surely have both higher accuracy and efficiency functionality for real-time uses.

\section{Future Work and Ideas}

In the future research, giving more attention to the deep leaning models through mechanisms can help them a lot to focus on the solar panel images main points more such as places with dust and faults accumulative. One has to proceed with complex mechanism such as self-attention or multi-head attention to catch attention on the vital parts.

The blend of multi-modular connection may be set up by inclusion of extra sensor feeds like infrared imaging or environment conditions, which will heighten the toughness and precision of the detection system. This method puts together information from diverse sources in order to come to a conclusion that covers all aspects regarding the working mechanics of the solar panel.

Transfer learning techniques may encourage model training and from scratch. We will adjust these models purposely on the solar panels' data to make additionally zone adaptation methods that we will use to improve the generalisation across different environments.

Using the methods focusing on uncertainty estimation would allow to better understand the weaknesses and strengths of the detection system. An indispensible part of everyday operations is to reach the level of confidence, which entails decision-makers being capable of judging the consequences of different scenarios and act thereupon with the obtained information.

The creation of online learning platforms with active learning will be the driving force for creating a new data based model ammendment. Online learning and active learning interactively update parameters on the go and smartly choose valuable samples for annotation, hence accelerates performance to the limit where human intervention is minimally involved.

Inferencing deep learning models for applications of real-time deployment on the edge computing devices makes it possible for the solar panels to be monitored on-site and maintenance actions to be taken promptly and with less downtime. As to edge deployment is the way to affix scalability but accessibility is even more concrete in the regions that are remote.

Expanding research into both long-term monitoring during the solar panel installation and also proactive maintenance can be helpful for detecting early warnings concerning material degradation or performance loss. Analyzing the past data provides schemes of proactive maintenance to replace the system when it is at the verge of breakdown.

\end{document}